\documentclass{bmvc2k}
\usepackage{enumitem}
\usepackage[ruled, linesnumbered]{algorithm2e}
\usepackage{booktabs}
\usepackage{multirow}
\usepackage{graphicx}
\usepackage{makecell}
\usepackage{tikz, pgfplots}
\usepackage{amssymb}
\usepackage{amsmath}

\title{Meta Transferring for Deblurring}

\addauthor{Po-Sheng Liu}{bensonliu.cs09@nycu.edu.tw}{1}
\addauthor{Fu-Jen Tsai}{fjtsai@gapp.nthu.edu.tw}{2}
\addauthor{Yan-Tsung Peng}{ytpeng@cs.nccu.edu.tw}{3}
\addauthor{Chung-Chi Tsai}{charles0184@gmail.com}{4}
\addauthor{Chia-Wen Lin}{cwlin@ee.nthu.edu.tw}{2}
\addauthor{Yen-Yu Lin}{lin@cs.nycu.edu.tw}{1}

\addinstitution{
 National Yang Ming Chiao Tung University, Taiwan\\
}
\addinstitution{
 National Tsing Hua University, Taiwan\\
}
\addinstitution{
 National Chengchi University, Taiwan\\
}
\addinstitution{
 Qualcomm Technologies, Inc., United States\\
}

\runninghead{Liu ET AL}{Meta Transferring for Deblurring}

\def\etal{\emph{et al}\bmvaOneDot}

\newcommand \footnoteonlytext[1]
{
    \let \mybackup \thefootnote
    \let \thefootnote \relax
    \footnotetext{#1}
    \let \thefootnote \mybackup
    \let \mybackup \imareallyundefinedcommand
}

\newcommand{\red}[1]{\textcolor{red}{#1}}
\renewcommand{\arraystretch}{1.1} 
\graphicspath{{''}}

\begin{document}
\maketitle

\begin{abstract}
%
%
%
%
%
%
%
Most previous deblurring methods were built with a generic model trained on blurred images and their sharp counterparts. However, these approaches might have sub-optimal deblurring results due to the domain gap between the training and test sets. This paper proposes a reblur-deblur meta-transferring scheme to realize test-time adaptation without using ground truth for dynamic scene deblurring. Since the ground truth is usually unavailable at inference time in a real-world scenario, we leverage the blurred input video to find and use relatively sharp patches as the pseudo ground truth. Furthermore, we propose a reblurring model to extract the homogenous blur from the blurred input and transfer it to the pseudo-sharps to obtain the corresponding pseudo-blurred patches for meta-learning and test-time adaptation with only a few gradient updates. Extensive experimental results show that our reblur-deblur meta-learning scheme can improve state-of-the-art deblurring models on the DVD, REDS, and RealBlur benchmark datasets.The source code is available at {\color{blue}\url{https://github.com/po-sheng/Meta_Transferring_for_Deblurring}}.
\end{abstract}

\section{Introduction}
\label{sec:intro}
\begin{figure}[t]
\includegraphics[width=\textwidth]{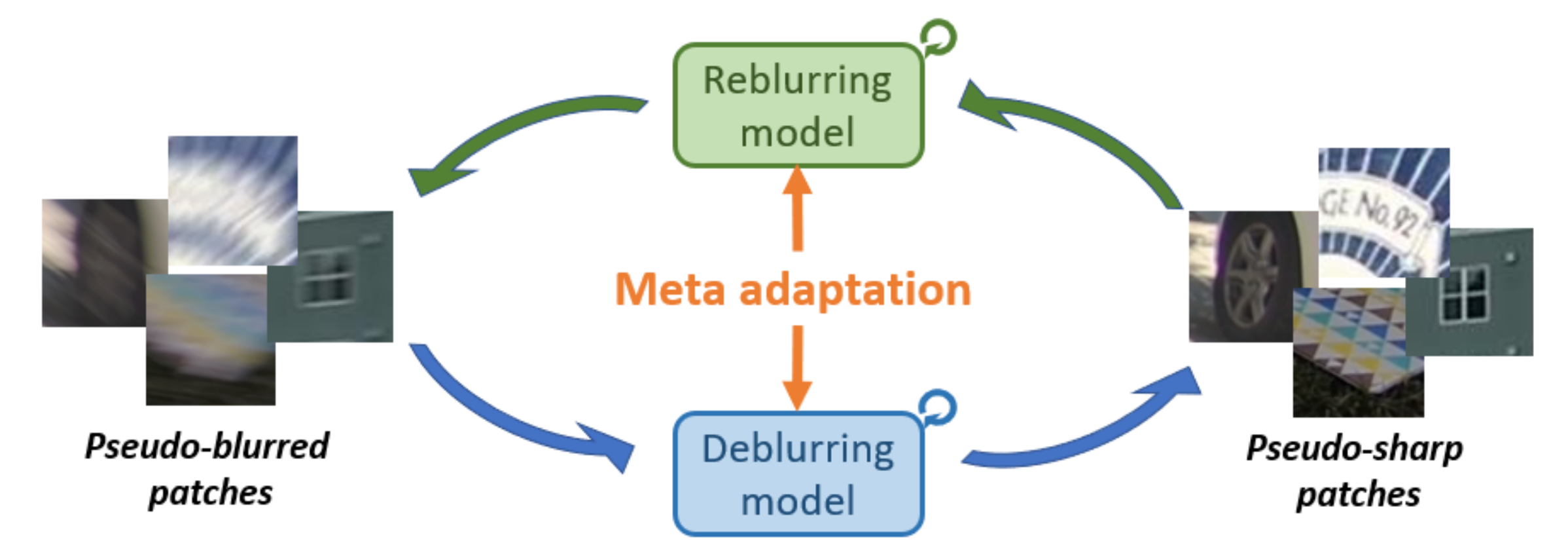}
\caption{Illustration of the proposed reblur-deblur meta-transferring scheme. We utilize the reblurring model to generate pseudo-blurred and pseudo-sharp pairs to facilitate meta-learning for the test-time adaptation. Note that pseudo-sharp patches are selected from blurred video frames.}
\label{fig:teaser}
\end{figure}
Dynamic scene deblurring aims to recover sharp images from blurred ones caused by camera shakes or moving objects. It is challenging to recover images with such blur since dynamic scene blur is often non-uniform and directional. Using methods with some prior assumptions~\cite{Pan_2016_cvpr,8100221,Chen_2019_CVPR} or blur uniformity~\cite{Cho_2009_ACM} can only achieve limited performance.
Existing learning-based methods for single image deblurring~\cite{Zamir2021MPRNet,park2020multi,zamir2021restormer,wang2021uformer,cho2021rethinking,kupyn2019deblurgan,zhang2019deep} or video deblurring~\cite{li2021arvo,deng2021multi,suin2021gated,zhong2020efficient,pan2020cascaded,argaw2021motion,lin2020learning,shen2020blurry,zhou2019spatio,nah2019recurrent} could obtain better results by training these models in a supervised manner, learning to deblur images from training with blurred and sharp image pairs. 
Although significant progress has been made using these supervised-learning methods, they did not consider exploiting rich internal information in test data and tend to be sub-optimal during testing.  
Therefore, we aim to build a domain adaptation strategy that allows test-time adaptation to improve existing deblurring methods. 
%
%
%
%
%
Few deblurring works have been done using meta-learning since ground truth is not accessible during testing. Chi \etal.~\cite{chi2021test} proposed a self-supervised meta-auxiliary learning strategy that meta-trains and meta-tests the model by restoring the input blurred image. However, it could only achieve sub-optimal adaptation since blur self-restoration may not help to deblur. 
%
%
By contrast, we observe that a scene appearing in multiple frames in a video often has different blurring degrees. Some local patches are relatively sharp, whereas some are relatively blurred. Therefore, we propose to leverage these patches to synthesize sharp and blurred pairs to enable meta-learning for dynamic scene deblurring during testing without access to the ground truth like~\cite{chi2021test} but in a different approach. 
%
More specifically, this work proposes a reblur-deblur meta-transferring scheme to generate pseudo-blurred and pseudo-sharp pairs to achieve test-time adaptation in meta-learning, as shown in Fig.~\ref{fig:teaser}.
%
%
%
Our reblurring model can transfer blurred patterns homogeneous to the task to pseudo-sharp patches selected to synthesize pseudo-blurred patches, which can serve as pseudo-sharp-and-blurred pairs as the support set in meta-learning.
%
%
%
%
Our contributions are three-fold. First, we propose a novel 
reblur-deblur meta-transferring scheme that can generate pseudo-blurred and pseudo-sharp pairs for meta-learning.
Second, the proposed scheme facilitates meta-learning for dynamic scene deblurring without extra training data needed.
Third, extensive experimental results show that our method improves the performance of existing deblurring models on various datasets, including DVD~\cite{su2017deep}, REDS~\cite{nah2019ntire}, and RealBlur-J~\cite{rim2020real}.

\section{Related Works}
\label{sec:related}
\subsection{Dynamic scene deblurring}
Previous dynamic scene deblurring methods can be roughly divided into single image deblurring~\cite{Zamir2021MPRNet,park2020multi,zamir2021restormer,wang2021uformer,cho2021rethinking,kupyn2019deblurgan,zhang2019deep,Tsai2022Stripformer, Tsai2022BANet} and video deblurring~\cite{li2021arvo,deng2021multi,suin2021gated,zhong2020efficient,pan2020cascaded,argaw2021motion,lin2020learning,shen2020blurry,zhou2019spatio,nah2019recurrent}. 
The main difference is the number of input frames. Video deblurring utilizes consecutive frames to help deblurring, which can obtain more information than image deblurring. 
However, most existing methods fail to consider internal information from testing data and ignore the problem of domain gap between training and testing sets, which would be sub-optimal during testing.
%

\subsection{Meta learning on low-level vision tasks}
Previous low-level vision methods commonly train a generic model using training data without considering the rich internal information in test data. 
%
%
Recently, a few methods developed for low-level vision tasks have been proposed~\cite{park2020fast,soh2020meta,chi2021test}. Super-resolution (SR) methods~\cite{park2020fast,soh2020meta} utilized MAML~\cite{finn2017model} to efficiently adapt to test images for better SR results, where they used the patch-recurrence property intrinsic to an image itself, downsampling images to generate low- and high-resolution training pairs during testing.
%
%
Since finding such an intrinsic characteristic from a blurred image is challenging, Chi \etal.~\cite{chi2021test} proposed a meta-auxiliary learning method, where they meta-trained the deblurring model to learn the input blurred image in a self-supervised manner to help deblur. Thus, it can enable the model to adapt to test blurred images without using their ground-truth counterparts.
Nevertheless, restoring the input blurred image and deblurring it essentially works in a contradictory fashion, which may only produce sub-optimal results. 
%
%
Unlike meta-auxiliary learning for deblurring proposed in~\cite{chi2021test} requires attaching additional layers to train the model, our work builds a reblurring model that can transfer blur patterns from relative-blurred to relative-sharp patches. It can synthesize pseudo-blurred and pseudo-sharp pairs for deblurring models without modifying them to facilitate meta-learning. 

\subsection{Reblurring for deblurring.}
Some recent deblurring works~\cite{chen2018reblur2deblur,nah2021clean,park2020blur,zhang2020deblurring} were proposed to generate additional blurred images to train the models. 
%
Zhang \etal.~\cite{zhang2020deblurring} used an extra real blurred dataset to produce synthetic blurred images using generative adversarial networks. 
%
Park \etal.~\cite{park2020blur} used a recurrent method to produce images with various blurring degrees. However, the generated data for training may not help upon testing. 
To enable test-time adaptation, Nah \etal.~\cite{nah2021clean} utilizes a reblurring loss to constrain deblurring results using a pre-trained reblurring model based on the observation that clean images are hard to reblur.
Chen \etal.~\cite{chen2018reblur2deblur} chose to self-supervise a model for deblurring, similar to~\cite{chi2021test}. 
In contrast, we leverage the internal information in testing videos to generate domain-aware pseudo-blurred and pseudo-sharp pairs, enabling meta-learning better than~\cite{chen2018reblur2deblur, chi2021test}.

%

\section{Proposed Method}
\label{sec:method}
\subsection{Overview}
\label{subsec::overview}
\vspace{-0.05in}

\begin{figure}[t]
\includegraphics[width=\textwidth]{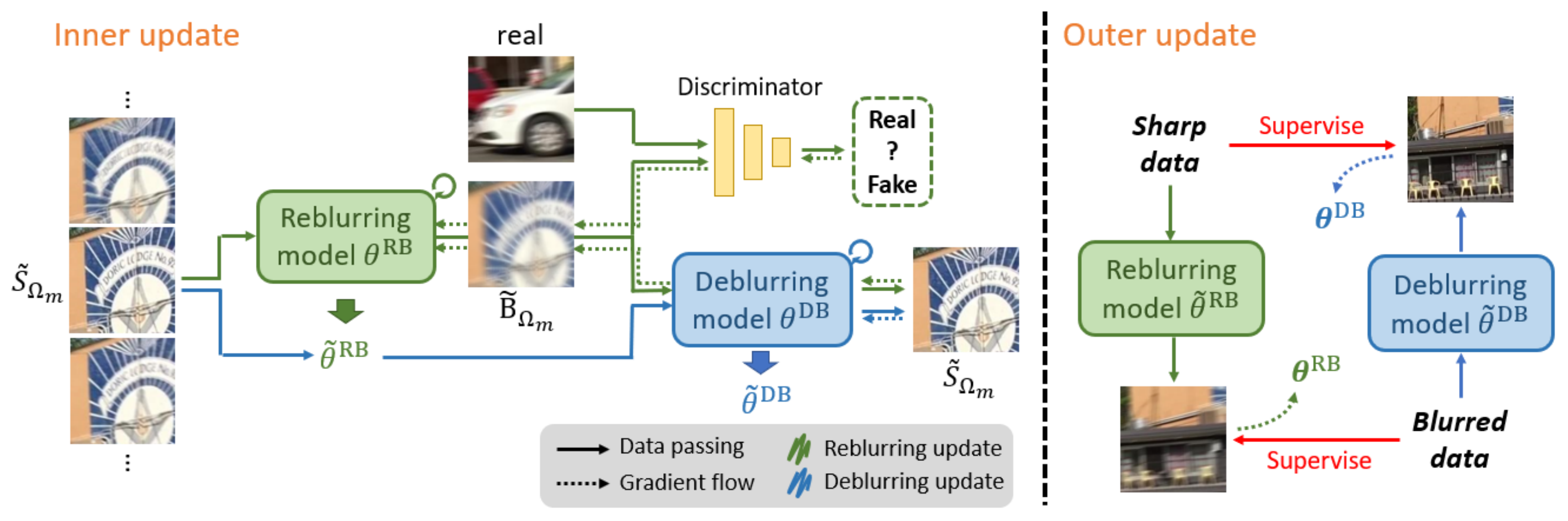}
\caption{
Illustration of our reblur-deblur meta-transferring training scheme. In the inner update, we use an adapted reblurring model to generate a domain-aware support set, enabling meta-learning to update the deblurring model. In the outer update, we evaluate the adapted models by query set to obtain meta-learned weights $\theta^{RB}$ and $\theta^{DB}$.
}
\label{fig:scheme}
\end{figure}

The paper proposes a reblur-deblur meta-transferring scheme for test-time adaptation, composed of meta-training and meta-testing phases. 

In meta-training, we consider deblurring each video $V_i$ in the training set as a task, where $V_i$ contains $N$ blurred frames $B_n$ and their sharp counterparts $S_n$ ($n=\{1, 2, ..., N\}$). Figure~\ref{fig:scheme} depicts the proposed reblur-deblur meta-transferring training scheme, where the inner update trains a deblurring model on the support set generated by the reblurring model, and the outer update trains the deblurring and reblurring models adapted to the query set. Like~\cite{chi2021test}, not needing the ground truth at inference as well, we instead use the proposed reblurring model to generate $M$ pseudo-blurred patches $\tilde{B}_{\Omega_m}$ from pseudo-sharp patches $\tilde{S}_{\Omega_m}$ selected from $\{{B_n}\}_{n=1}^N$. Based on our observation that a scene often has various blurring degrees in multiple frames in a blurred video, we can pick relatively sharp patches in blurred frames to be pseudo-sharp patches. To choose sharp patches, we propose to measure the blurring degree by a self-shift method, and those with the least blur are selected. These $M$ pseudo-blurred and pseudo-sharp pairs $\{\tilde{B}_{\Omega_m}, \tilde{S}_{\Omega_m}\}_{m=1}^M$ are used as the support set. For the query set, we use the whole $V_i = \{{B_n}, {S_n}\}_{n=1}^N$ to train the reblurring and deblurring models. 

In meta-testing, given any testing video $V^{test}_i$, we generate the support set as in the meta-training phase, not needing the ground truth. Next, we explain how to create the support set for meta-learning in Sec.~\ref{subsec::Support Set} and the architecture of the reblurring model in Sec.~\ref{subsec::intro_arch}. At last, Sec.~\ref{subsec::intro_scheme} details the proposed reblur-deblur meta-transferring scheme for test-time adaptation. 

\subsection{Generation of the Support Set}
\label{subsec::Support Set}
%
%
%
To enable meta-learning for the test-time adaptation, we generate pseudo-blurred and pseudo-sharp pairs $\{\tilde{B}_{\Omega_m}, \tilde{S}_{\Omega_m}\}_{m=1}^M$ from each blurred video. 
These pseudo-sharp patches $\tilde{S}_{\Omega_m}$ are fed into a reblurring model to generate pseudo-blurred patches $\tilde{B}_{\Omega_m}$. The support set contains $\{\tilde{B}_{\Omega_m}, \tilde{S}_{\Omega_m}\}_{m=1}^M$.

\begin{figure}[t]
\includegraphics[width=\textwidth]{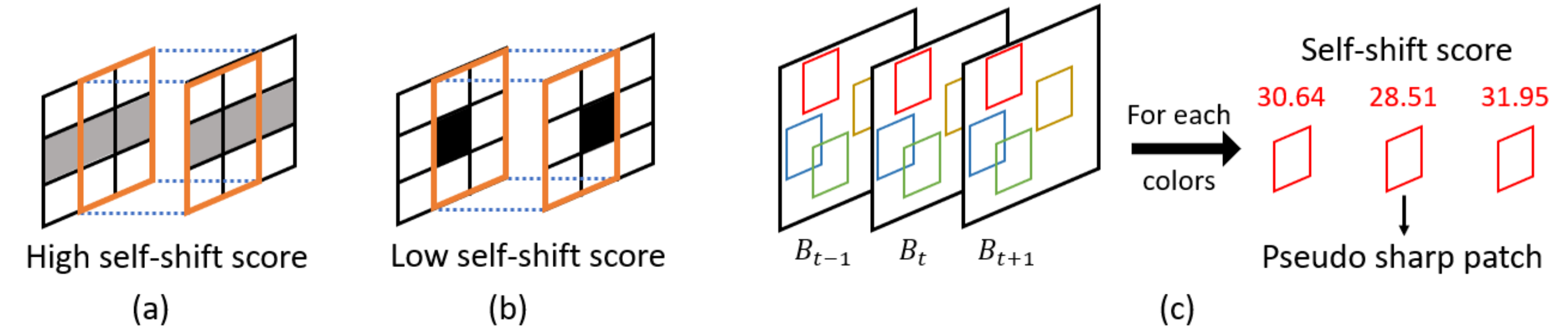}
\caption{The proposed self-shift method. The self-shift compares the shifted image and its original input on PSNR. (a) A high self-shift score implies blur. (b) A low-shift score implies sharpness. (c) We select pseudo-sharp patches from blurred video frames with low self-shift scores.}
\label{fig:ss}
\end{figure}

%
%
%
%
%
%
%
%
%
To search for relatively sharp patches in a blurred video, we propose to measure the blurring degree by a self-shift method. Based on our observation, sharp images usually have stronger edges with larger gradients than blurred images. Fig.~\ref{fig:ss} illustrates the self-shift method, where four images are generated by shifting the input image one pixel to the right, top, top-right, and top-bottom, in four directions. 
We obtain a self-shift score by averaging four PSNR values calculated by the four shifted images with the input image as the reference. 
A higher score indicates that the shifted images are similar to their original image (Fig.~\ref{fig:ss} (a)), implying it is blurred. In comparison, a lower score represents that the input image is sharp. 
We randomly select $M$ locations to crop image patches, where the relatively sharp patches are chosen with the lowest shift scores among all the colocated patches in all the input video frames, shown in Fig.~\ref{fig:ss} (c).
Finally, we feed $\{\tilde{S}_{\Omega_m}\}_{m=1}^M$ into the reblurring model to generate pseudo-blurred patches $\{\tilde{B}_{\Omega_m}\}_{m=1}^M$ to generate the support set.

\subsection{Reblurring Model}
\label{subsec::intro_arch}
\begin{figure}[t]
\includegraphics[width=\textwidth]{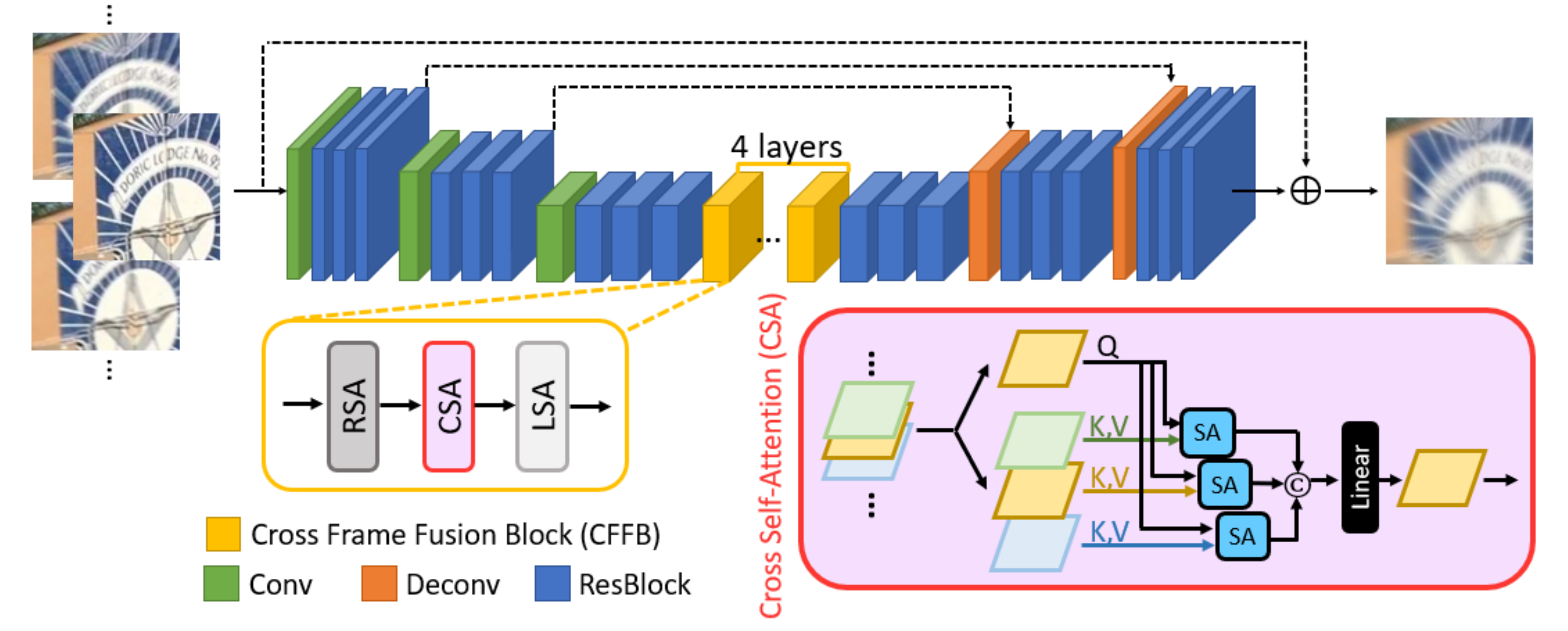}
\caption{We use cross self-attentions to transfer the extracted blurred pattern from the collocated patches in the neighboring frames of the input pseudo-sharp patch to generate a domain-aware pseudo-blurred and sharp pair.}
\label{fig:arch}
\end{figure}

%
%
%
Fig.~\ref{fig:arch} shows the architecture of the proposed video reblurring network.  For each pseudo-sharp patch $\tilde{S}_{\Omega_m}$, we choose collocated patches in the five frames neighbored by the sharp patch $\{\tilde{S}_{\Omega_{m,t-2}}, ...,\tilde{S}_{\Omega_{m,t+2}}\}$, where $\tilde{S}_{\Omega_{m,t}}=\tilde{S}_{\Omega_m}$, and $t$ denotes the frame index, to generate the corresponding pseudo-blurred patch $\tilde{B}_{\Omega_m}$. 
In the reblurring model, the five colocated patches $\{\tilde{S}_{\Omega_{m,t-2}}, ...,\tilde{S}_{\Omega_{m,t+2}}\}\in\mathbb{R}^{T\times H\times W\times 3}$ are fed into an encoder to generate embedding features $\mathbf{F}\in\mathbb{R}^{T\times H'\times W'\times C}$ , where $T$, $H$, $H'$, $W$, $W'$, and $C$ denote the number of patches ($T=5$ here), height, embedded feature height, width, embedded feature width, and channel dimensions, respectively.
Inspired by~\cite{chen2021regionvit}, we propose a cross-frame fusion block (CFFB), containing regional self-attention (RSA), cross self-attention (CSA), and local self-attention (LSA)~\cite{chen2021regionvit}.

The CFFB divides $\mathbf{F}$ into non-overlapping regional tokens with a window size of $r$ as $\mathbf{F}\in\mathbb{R}^{T\times N \times D}$, where $N=\frac{H'W'}{r^2}$ and $D=C\times{r^2}$ denote the number of tokens and dimensions.
To reduce the channel dimensions, we use an embedding layer to embed $\mathbf{F}$ into $\mathbf{F}\in\mathbb{R}^{T\times N \times D^{\prime}}$, where $D^{\prime}=2C$. 
We then feed $\mathbf{F}$ into RSA~\cite{chen2021regionvit}, which performs self-attention on regional tokens to generate $\mathbf{F}^{RSA}\in\mathbb{R}^{T\times N \times D^{\prime}}$, where $T$ is regarded as the number of batches. 
To utilize cross-frame information, we propose CSA, which considers the center frame patch tokens $F^{RSA}_t\in\mathbb{R}^{N \times D^{\prime}}$ from $\mathbf{F}^{RSA}\in\{F^{RSA}_{t-2}, ...,F^{RSA}_{t+2}\}$ as the query to perform self-attention with $F^{RSA}_{t^{\prime}}\in\mathbb{R}^{N \times D^{\prime}}$, where $t^{\prime}\in \{t-2, ..., t+2\}$, i.e.,
\begin{equation}    
    \begin{gathered}
    Q = F_{t}^{RSA}W^q; \ \ 
    K_{t^{\prime}} = F_{t^{\prime}}^{RSA}W^k; \ \ 
    V_{t^{\prime}} = F_{t^{\prime}}^{RSA}W^v; \\
    F^{CSA}_{t^{\prime}} = \mathbf{Softmax}(\frac{Q(K_{t^{\prime}})^T}{\sqrt{D^{\prime}}})V_{t^{\prime}},
    \end{gathered}
\label{eq:csa}
\end{equation}
where $W^q$, $W^k$, and $W^v$ $\in \mathbb{R}^{D^{\prime}\times D^{\prime}}$ denote the embedding weights of the query, key, and value.
We concatenate the output features followed by a linear layer with dimensions $\mathbb{R}^{5D^{\prime}\times D^{\prime}}$ to generate $ \tilde{F}^{CSA}_t\in \mathbb{R}^{N\times D^{\prime}}$ as
%
\begin{equation}    
    \tilde{F}^{CSA}_t = \mathbf{Linear}(\mathbf{Concat}([F^{CSA}_{t-2},\ F^{CSA}_{t-1},\ F^{CSA}_{t},\ F^{CSA}_{t+1},\ F^{CSA}_{t+2}])).
\label{eq:linear_cat}
\end{equation}
After CSA, we replace the $F_t^{RSA}$ in $\mathbf{F}^{RSA}$ with $\tilde{F}^{CSA}_t$ to obtain a new $\mathbf{F}^{RSA}\in\mathbb{R}^{T\times N \times D^{\prime}}$, where $\tilde{F}^{CSA}_t$ has fused the neighboring information.
Lastly, we feed $\mathbf{F^{RSA}}$ and initial input features $\mathbf{F}$ into LSA to obtain $\mathbf{F}^{LSA}\in\mathbb{R}^{T\times N \times D^{\prime}}$ like~\cite{chen2021regionvit}.
After four CFFBs, the middle frame patch tokens $F_{t}^{LSA}\in\mathbb{R}^{N \times D^{\prime}}$ are fed into a linear layer to increase channel dimensions from $D^{\prime}$ to $D$ and folded into $F^{O}\in\mathbb{R}^{H'\times W' \times C}$ followed by a decoder to generate the pseudo-blurred patch $\tilde{B}_{\Omega_m}\in\mathbb{R}^{H\times W \times 3}$.

The reblurring model needs to be pre-trained, where we use blurred and sharp image pairs in the training set with the Charbonnier loss~\cite{Lai_2017_CVPR} and adversarial loss~\cite{Wgan} derived from an addional discriminator, i.e., five consecutive sharp frames as input and the middle blurred counterpart as the ground truth. 
%
%
%
%
%
To address the domain gap of the pre-trained reblurring model, we also update the reblurring model and discriminator during testing.
Therefore, we generate realistic pseudo pairs $\{\tilde{B}_{\Omega_m}, \tilde{S}_{\Omega_m}\}_{m=1}^M$ as the support set by transferring blurred patterns at inference to enable meta-learning.

\subsection{Reblur-Deblur Meta-Transferring}
\label{subsec::intro_scheme}
Here, we introduce the meta-learning process for the reblurring and deblurring models. 

\noindent\textbf{Meta training:}
We consider deblurring each video $V_i$ in the training set as a task, where $V_i = \{{B_n}, {S_n}\}_{n=1}^N$ are available during training.
%
We initialize the reblurring model (parameterized by $\theta^{RB}$), discriminator ($\theta^{D}$), and deblurring model ($\theta^{DB}$) with their pre-trained weights.
For each video $V_i$, we first initialize the reblurring, discriminator, and deblurring weights $\tilde{\theta}^{RB}_{i}$, $\tilde{\theta}^{D}_{i}$, and $\tilde{\theta}^{DB}_{i}$ by $\theta^{RB}$, $\theta^{D}$, and $\theta^{DB}$, respectively.
As shown in Fig.~\ref{fig:scheme},  the meta-process can be split into two stages: one is the inner update to obtain the reblurring and deblurring models using the support set, and the other is the outer update to evaluate the ability of adapted models by the query set. 

In the inner update, we first update the reblurring model to generate the support set for the deblurring model. 
We choose $M$ pseudo-sharp patches $\{\tilde{S}_{\Omega_m}\}_{m=1}^M$ from $\{{B_n}\}_{n=1}^N$ by the self-shift method to generate pseudo-blurred patches $\{\tilde{B}_{\Omega_m}\}_{m=1}^M$ from the reblurring model $f^{RB}(\cdot)$ parameterized by $\tilde{\theta}^{RB}_{i}$ as
\begin{equation}    
\label{reblur_model}
    \tilde{B}_{\Omega_m} = f^{RB}(\{\tilde{S}_{\Omega_{m,t-2}}, ...,\tilde{S}_{\Omega_{m,t+2}}\}; \tilde{\theta}^{RB}_{i}),
\end{equation}
where the five consecutive frame patches from the pseudo-sharp patch are concatenated as the input.
%
To generate realistic and homogenous blurred patterns, we use the discriminator parameterized by $\tilde{\theta}^{D}_{i}$ to calculate the adversarial loss $\mathcal{L}_{adv}$. Here, the pseudo-blurred patch $\tilde{B}_{\Omega_m}$ is considered a faked blurred image, and we select a real blurred patch by the self-shift method with a high score. 
Additionally, we add a cycle loss based on cycle consistency from the deblurring model $f^{DB}(\cdot)$ parameterized by ${\theta}^{DB}$ as 
\begin{equation}    
\label{eq:cycle}
    \mathcal{L}_{cycle} =  \mathcal{L}(f^{DB}(\tilde{B}_{\Omega_m};\theta^{DB}),\ \tilde{S}_{\Omega_m}), 
\end{equation}
where $\mathcal{L}$ is the loss function adopted in the deblurring model. It represents that deblurring the generated pseudo-blurred patch by the reblurring model should obtain its corresponding pseudo-sharp.
For updating the reblurring model, we first update the discriminator $\tilde{\theta}^{D}_{i}$ by $\mathcal{L}_{adv}$ 
%
and then update the reblurring model by the total reblurring loss as $\mathcal{L}_{RB} = \mathcal{L}_{adv} + \lambda\mathcal{L}_{cycle}$, where we set $\lambda= 0.01$ to focus more on the blur-transferring ability of the disciminator, i.e.,
\begin{equation}    
\label{reblur_update}
    \tilde{\theta}^{RB}_{i}=\tilde{\theta}^{RB}_{i}-\alpha\nabla_{\tilde{\theta}^{RB}_{i}} (\mathcal{L}_{RB}),
\end{equation}
where we iteratively update the  $\tilde{\theta}^{D}_{i}$ and $\tilde{\theta}^{RB}_{i}$ by the $M$ pseudo pairs with the learning rate of $\alpha$. 
At last, the reblurring model ($\tilde{\theta}^{RB}_{i}$) can effectively transfer blurred patterns extracted from the test domain to generate realistic pseudo-blurred patches.

After the adaptation of reblurring model $\tilde{\theta}^{RB}_{i}$, we generate the support set $\{\tilde{B}_{\Omega_m}, \tilde{S}_{\Omega_m}\}_{m=1}^M$ by Eq.~\ref{reblur_model} and iteratively update the deblurring model using these $M$ pairs as
\begin{equation}    
\label{deblur_update}
    \tilde{\theta}^{DB}_{i}=\tilde{\theta}^{DB}_{i}-\beta\nabla_{\tilde{\theta}^{DB}_{i}} (\mathcal{L}(f^{DB}(\tilde{B}_{\Omega_m};\tilde{\theta}^{DB}_{i}),\ \tilde{S}_{\Omega_m})),
\end{equation}
where $\mathcal{L}$ is the loss function used by the deblurring method, and $\beta$ is its learning rate.

In the outer update, we verify the adaptability of adapted models for each video $V_i$. 
We utilize all the blurred and sharp images $\{{B_n}, {S_n}\}_{n=1}^N$ in $V_i$ as the query set to perform meta-update based on the gradient calculated using the query set as
\begin{gather}
    \theta^{RB} = \theta^{RB} - \alpha \nabla_{\theta^{RB}} \sum_{n=1}^{N}  \mathcal{L}_{char}(f^{RB}(S_n;\tilde{\theta}^{RB}_{i}),\ B_n), \notag\\
    \theta^{DB} = \theta^{DB} - \beta \nabla_{\theta^{DB}} \sum_{n=1}^{N}  \mathcal{L}(f^{DB}(B_n;\tilde{\theta}^{DB}_{i}),\ S_n), 
\label{meta_update}
\end{gather}
where $L_{char}$ is the Charbonnier loss~\cite{Lai_2017_CVPR}. Since we regard deblurring on each video ${V_i}$ as a task, 
the meta-training process is repeated on all the available training videos until the model converges.
Through Eq.~\ref{meta_update}, we can obtain meta-learned weights $\theta^{RB}$ and $\theta^{DB}$, more transferable to a new task (video) compared to the original pre-trained weights.

\noindent\textbf{Meta testing:} Given any testing video $V^{test}_i = \{{B_n}\}_{n=1}^N$, we take meta-learned reblurring ($\theta^{RB}$), deblurring ($\theta^{DB}$) weights and pre-trained discriminator ($\theta^{D}$) weights as initialization.
Similar to the meta-training phase, we choose $M$ pseudo-sharp patches $\{\tilde{S}_{\Omega_m}\}_{m=1}^M$ from $\{{B_n}\}_{n=1}^N$ by the self-shift method and iteratively update the reblurring model using Eq.~\ref{reblur_update} to obtain the adapted reblurring model ($\tilde{\theta}^{RB}_{test}$).  
Next, we use the reblurring model ($\tilde{\theta}^{RB}_{test}$) to generate the support set to update the deblurring model by Eq.~\ref{deblur_update}.
Lastly, we deblur the whole testing video $V^{test}_i$ by the adapted deblurring model ($\tilde{\theta}^{DB}_{test}$). 
Because our support set is generated by pseudo-blurred and pseudo-sharp patches $\{\tilde{B}_{\Omega_m}, \tilde{S}_{\Omega_m}\}_{m=1}^M$, we can enable meta-learning on the deblurring task without the ground truth.
The detailed training and testing processes are shown in the supplementary.
\section{Experiments}
\label{sec:exp}
In this section, we evaluate the effectiveness of the proposed method. We first explain datasets and implementation details in Sec.~\ref{exp:detail}. 
We then demonstrate the quantitative and qualitative results in Sec.~\ref{exp:result}. 
In the end, we conduct ablation studies in Sec.~\ref{exp:ablation}.
\footnoteonlytext{The authors from the universities in Taiwan completed the experiments on the datasets.}

\subsection{Datasets and Implementation Details}
\label{exp:detail}
We trained all the compared models on the GoPro~\cite{nah2017deep} training set, containing $22$ videos, totally having $2,103$ training blurred and sharp image pairs in both the pre-training and meta-training stages.
At the meta-testing time, we use three datasets to demonstrate the effectiveness of the proposed test-time adaptation scheme, including DVD~\cite{su2017deep} testing set (10 videos, having 1,000 images), REDS~\cite{nah2019ntire} validation set (30 videos, having 3,000 images), and RealBlur-J~\cite{rim2020real} testing set (50 videos, having 980 images).
%
First, we pre-trained the reblurring model on the GoPro training set with a batch size of eight for $1,000$ epochs by Adam optimizer~\cite{kingma2014adam}, where the initial learning rate is set to $10^{-4}$, and the decay is $10^{-8}$ with the cosine annealing strategy.
%
We adopted random cropping, flipping, and rotating for data augmentation. 
%
Next, for meta-training, we randomly cropped $256 \times 256$ patches for the support and query sets. Note that the support set contains pseudo-sharp patches selected from these cropped patches and then pseudo-blurred patches generated by the pseudo-sharps using the reblurring model. 
%
We set the learning rate $\alpha$ to $10^{-6}$ in the reblurring model and $\beta$ to $2.5\times 10^{-6}$ in the deblurring model.
All methods are tested on an Nvidia A5000 GPU.

\setlength{\tabcolsep}{5pt} 
\renewcommand{\arraystretch}{1} 
\begin{table}[t!]
\centering
\caption{Evaluation results on three datasets and four SOTA deblurring models. ``Baseline" means the deblurring results obtained using the original models pre-trained on GoPro. ``Meta" means the results using ``Baseline" with our reblur-deblur meta-transferring scheme.}
\begin{tabular}{ c c | c c c c c c}
    \toprule
    & & \multicolumn{2}{c}{\bf{DVD}} & \multicolumn{2}{c}{\bf{REDS}} & \multicolumn{2}{c}{\bf{RealBlur-J}}\\
    \cmidrule(lr){3-4} \cmidrule(lr){5-6} \cmidrule(lr){7-8}
    & & PSNR & SSIM & PSNR & SSIM & PSNR & SSIM \\
    \hline\hline
    \multirow{2}{*}{\bf{MIMOUNet+~\cite{cho2021rethinking}}} 
                          & Baseline & 29.43 & 0.914 & 26.43 & 
                            0.859 & 27.63 & 0.837 \\
                          & Meta & \bf{29.70} &  \bf{0.917} & \bf{26.73} & 0.859 & \bf{28.11} & \bf{0.851} \\
    \hline\hline
    \multirow{2}{*}{\bf{MPRNet~\cite{Zamir2021MPRNet}}} 
                          & Baseline & 29.68 & 0.918 & 26.85 & 
                            0.864 & 28.70 & 0.873 \\
                          & Meta & \bf{30.04} &  \bf{0.921} & \bf{27.05} & 0.864 & \bf{28.75} & \bf{0.876} \\
    \hline\hline
    \multirow{2}{*}{\bf{Restormer~\cite{zamir2021restormer}}} 
                          & Baseline & 29.67 & 0.916 & 26.93 & 
                            0.867 & 28.96 & 0.879 \\
                          & Meta & \bf{30.01} &  \bf{0.921} & \bf{27.21} & 0.867 & \bf{29.07} & \bf{0.885} \\
    \hline\hline
    \multirow{2}{*}{\bf{CDVD-TSP~\cite{pan2020cascaded}}} 
                          & Baseline & 30.86 & 0.938 & 27.17 &     
                            0.891 & 28.69 & 0.873 \\
                          & Meta & \bf{30.97} &  \bf{0.939} & \bf{27.52} & \bf{0.892} & \bf{29.15} & \bf{0.887} \\
    
    \bottomrule
\end{tabular}
\label{table:result}
\end{table}

\subsection{Experimental Results}
\label{exp:result}
\textbf{Quantitative Analysis:}
We employed our scheme on four different state-of-the-art deblurring models, including MIMOUnet+~\cite{cho2021rethinking}, MPRNet~\cite{Zamir2021MPRNet}, Restormer~\cite{zamir2021restormer}, and CDVD-TSP~\cite{pan2020cascaded}. The first three models are for image deblurring, and the last is for video deblurring. 
In Table~\ref{table:result}, we compare the results obtained with or without the proposed reblur-deblur meta-transferring scheme.
As can be seen, our scheme can improve existing deblurring methods by 0.27dB, 0.29dB, and 0.28dB in PSNR on average on the DVD, REDS, and RealBlur-J datasets, respectively, indicating that it achieves the test-time adaptation for better performance. 
%
%
%
%
%

\begin{figure}[t]
\includegraphics[width=\textwidth]{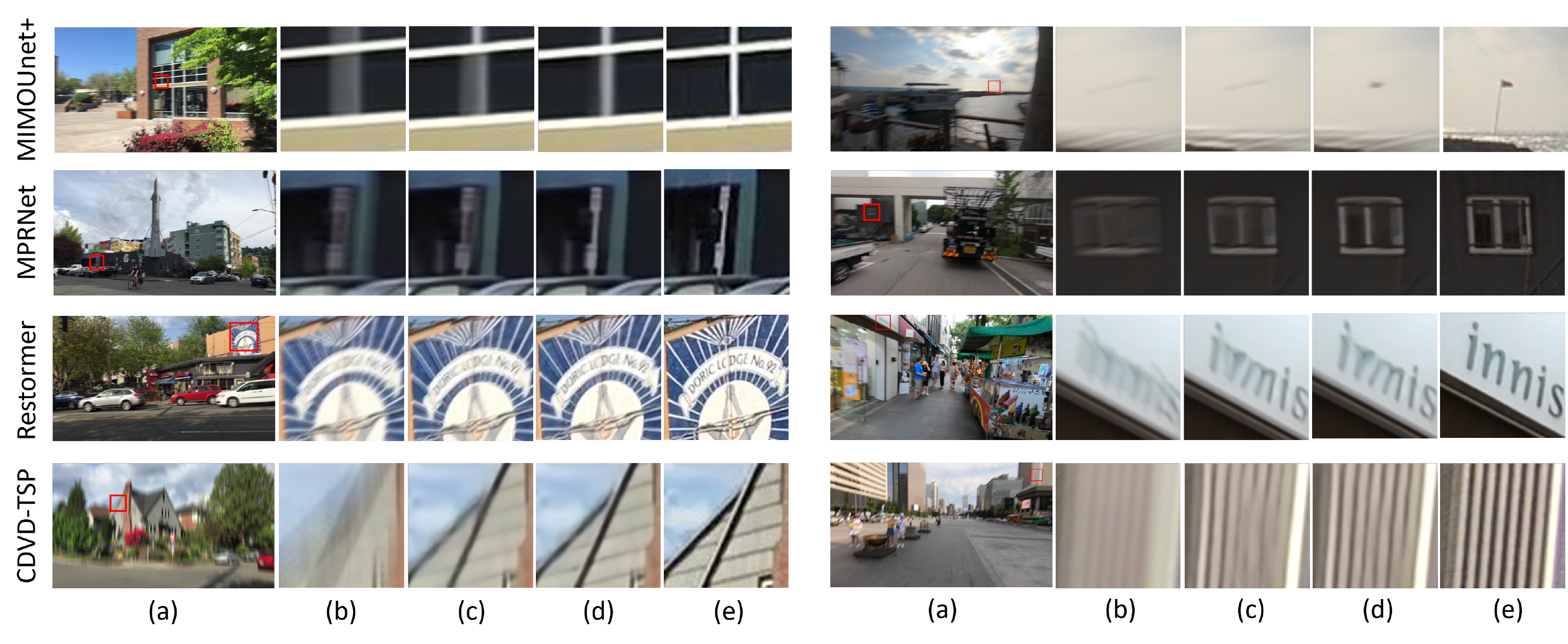}
\caption{Qualitative comparisons of deblurring results on DVD (left) and REDS (right) datasets. (a) Blurred input. (b) A zoom-in blurred patch, and its deblurred results using (c) Baseline and (d) Baseline /w our meta-transferring scheme. (e) Ground truth.}
\label{fig:visual}
\end{figure}

\noindent\textbf{Qualitative Analysis:} 
Fig.~\ref{fig:visual} shows visual comparisons of results obtained using the original state-of-the-art methods with or without using the proposed reblur-deblur meta-transferring scheme on public benchmark datasets. As shown, using our meta-transferring scheme restores the blurred content better, indicating that the proposed approach synthesizing images with homogenous blur to each task at inference time for test-time adaptation can effectively boost the deblurring performance. 
\subsection{Ablation Studies}
\label{exp:ablation}
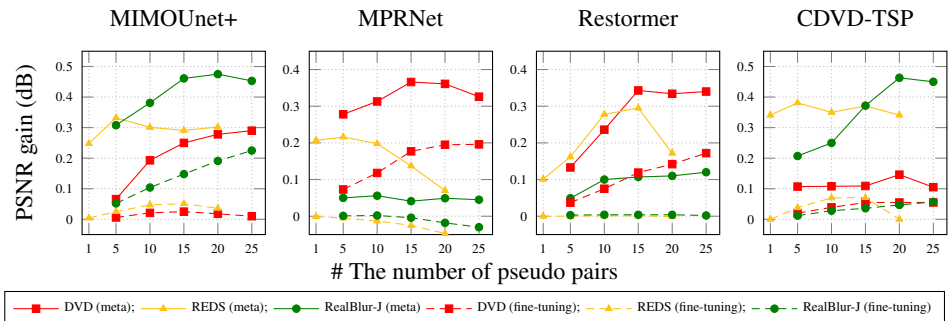
\begin{figure}[t!]
\centering
\begin{tabular}{ c }

\pgfplotsset{
    title style={font=\small},
    width=4cm, height=4cm,
    ylabel near ticks,
    label style={font=\small},
    tick label style={font=\tiny},
    ymin=-0.05 , ymax=0.55,
    ytick={0, 0.1, 0.2, 0.3, 0.4, 0.5},
    legend style={font=\tiny},
    grid=major,
    grid style=densely dotted,
    legend columns=-1,
    legend entries={DVD\ (meta);, REDS\ (meta);, RealBlur-J\ (meta), DVD\ (fine-tuning);, REDS\ (fine-tuning);, RealBlur-J\ (fine-tuning)},
    legend to name=Legends,
    legend style={text=black}
}

\begin{tikzpicture}
\begin{axis}[
    title={MIMOUnet+},
    ylabel={PSNR gain (dB)},
    xtick={1, 5, 10, 15, 20, 25},
    xmin=0 , xmax=27,
]
\addplot[
    color=red, 
    mark=square*,
    mark options={scale=0.7,fill=red}
    ]
    coordinates {(5, 0.066)(10, 0.193)(15, 0.250)(20, 0.278)(25, 0.29)};
\addplot[
    color=yellow!50!orange, 
    mark=triangle*,
    mark options={scale=0.7,fill=yellow!50!orange}
    ]
    coordinates {(1, 0.248)(5, 0.332)(10, 0.301)(15, 0.291)(20, 0.302)};
\addplot[
    color=green!50!black, 
    mark=*,
    mark options={scale=0.7,fill=green!50!black}
    ]
    coordinates {(5, 0.308)(10, 0.381)(15, 0.461)(20, 0.475)(25, 0.453)};
\addplot[
    color=red, 
    mark=square*,
    style=densely dashed,
    mark options={scale=0.7,solid,fill=red}
    ]
    coordinates {(5, 0.006)(10, 0.021)(15, 0.025)(20, 0.018)(25, 0.01)};
\addplot[
    color=yellow!50!orange, 
    mark=triangle*,
    style=densely dashed,
    mark options={scale=0.7,solid,fill=yellow!50!orange}
    ]
    coordinates {(1, 0.004)(5, 0.025)(10, 0.047)(15, 0.051)(20, 0.036)};
\addplot[
    color=green!50!black, 
    mark=*,
    style=densely dashed,
    mark options={scale=0.7,solid,fill=green!50!black}
    ]
    coordinates {(5, 0.052)(10, 0.104)(15, 0.148)(20, 0.191)(25, 0.225)};
\end{axis} 
\end{tikzpicture} 

\begin{tikzpicture}
\begin{axis}[
    title={MPRNet},
    ymin=-0.05 , ymax=0.45,
    ytick={0, 0.1, 0.2, 0.3, 0.4},
    xmin=0 , xmax=27,
    xtick={1, 5, 10, 15, 20, 25},
]
\addplot[
    color=red, 
    mark=square*,
    mark options={scale=0.7,fill=red}
    ]
    coordinates {(5, 0.278)(10, 0.313)(15, 0.366)(20, 0.361)(25, 0.326)};
\addplot[
    color=yellow!50!orange, 
    mark=triangle*,
    mark options={scale=0.7,fill=yellow!50!orange}
    ]
    coordinates {(1, 0.207)(5, 0.216)(10, 0.198)(15, 0.137)(20, 0.07)};
\addplot[
    color=green!50!black, 
    mark=*,
    mark options={scale=0.7,fill=green!50!black}
    ]
    coordinates {(5, 0.05)(10, 0.056)(15, 0.041)(20, 0.049)(25, 0.045)};
\addplot[
    color=red, 
    mark=square*,
    style=densely dashed,
    mark options={scale=0.7,solid,fill=red}
    ]
    coordinates {(5, 0.073)(10, 0.118)(15, 0.177)(20, 0.195)(25, 0.196)};
\addplot[
    color=yellow!50!orange, 
    mark=triangle*,
    style=densely dashed,
    mark options={scale=0.7,solid,fill=yellow!50!orange}
    ]
    coordinates {(1, -0.001)(5, -0.005)(10, -0.013)(15, -0.025)(20, -0.047)};
\addplot[
    color=green!50!black, 
    mark=*,
    style=densely dashed,
    mark options={scale=0.7,solid,fill=green!50!black}
    ]
    coordinates {(5, 0.001)(10, 0.002)(15, -0.004)(20, -0.018)(25, -0.03)};
\end{axis} 
\end{tikzpicture} 

\begin{tikzpicture}
\begin{axis}[
    title={Restormer},
    ymin=-0.05 , ymax=0.45,
    ytick={0, 0.1, 0.2, 0.3, 0.4},
    xmin=0 , xmax=27,
    xtick={1, 5, 10, 15, 20, 25},
]
\addplot[
    color=red, 
    mark=square*,
    mark options={scale=0.7,fill=red}
    ]
    coordinates {(5, 0.133)(10, 0.236)(15, 0.343)(20, 0.334)(25, 0.340)};
\addplot[
    color=yellow!50!orange, 
    mark=triangle*,
    mark options={scale=0.7,fill=yellow!50!orange}
    ]
    coordinates {(1, 0.101)(5, 0.162)(10, 0.278)(15, 0.295)(20, 0.172)};
\addplot[
    color=green!50!black, 
    mark=*,
    mark options={scale=0.7,fill=green!50!black}
    ]
    coordinates {(5, 0.049)(10, 0.1)(15, 0.107)(20, 0.11)(25, 0.12)};
\addplot[
    color=red, 
    mark=square*,
    style=densely dashed,
    mark options={scale=0.7,solid,fill=red}
    ]
    coordinates {(5, 0.037)(10, 0.075)(15, 0.119)(20, 0.142)(25, 0.172)};
\addplot[
    color=yellow!50!orange, 
    mark=triangle*,
    style=densely dashed,
    mark options={scale=0.7,solid,fill=yellow!50!orange}
    ]
    coordinates {(1, 0)(5, 0)(10, 0.001)(15, 0.001)(20, -0.001)};
\addplot[
    color=green!50!black, 
    mark=*,
    style=densely dashed,
    mark options={scale=0.7,solid,fill=green!50!black}
    ]
    coordinates {(5, 0.003)(10, 0.004)(15, 0.004)(20, 0.004)(25, 0.002)};
\end{axis} 
\end{tikzpicture} 

\begin{tikzpicture}
\begin{axis}[
    title={CDVD-TSP},
    xmin=0 , xmax=27,
    xtick={1, 5, 10, 15, 20, 25},
]
\addplot[
    color=red, 
    mark=square*,
    mark options={scale=0.7,fill=red}
    ]
    coordinates {(5, 0.107)(10, 0.108)(15, 0.109)(20, 0.146)(25, 0.105)};
\addplot[
    color=yellow!50!orange, 
    mark=triangle*,
    mark options={scale=0.7,fill=yellow!50!orange}
    ]
    coordinates {(1, 0.341)(5, 0.381)(10, 0.350)(15, 0.371)(20, 0.341)};
\addplot[
    color=green!50!black, 
    mark=*,
    mark options={scale=0.7,fill=green!50!black}
    ]
    coordinates {(5, 0.207)(10, 0.25)(15, 0.372)(20, 0.463)(25, 0.450)};
\addplot[
    color=red, 
    mark=square*,
    style=densely dashed,
    mark options={scale=0.7,solid,fill=red}
    ]
    coordinates {(5, 0.02)(10, 0.039)(15, 0.055)(20, 0.055)(25, 0.055)};
\addplot[
    color=yellow!50!orange, 
    mark=triangle*,
    style=densely dashed,
    mark options={scale=0.7,solid,fill=yellow!50!orange}
    ]
    coordinates {(1, 0)(5, 0.038)(10, 0.07)(15, 0.071)(20, 0)};
\addplot[
    color=green!50!black, 
    mark=*,
    style=densely dashed,
    mark options={scale=0.7,solid,fill=green!50!black}
    ]
    coordinates {(5, 0.012)(10, 0.028)(15, 0.036)(20, 0.047)(25, 0.057)};
\end{axis} 
\end{tikzpicture} \\

\small{\# The number of pseudo pairs} \\ \\ \\ 
\ref{Legends} \\

\end{tabular}
\caption{Ablations on meta-transferring vs. fine-tuning and analyses on different numbers of pseudo pairs included in the support set. The x-axis shows the number of pseudo pairs, and the y-axis shows the PSNR gain. Solid and dashed curves represent meta-testing and fine-tuning.}
\label{fig:sup_num}
\end{figure}

%
%
This section analyzes the ablation studies we conducted for the proposed approach. Table~\ref{table:reblur_loss} shows the effectiveness of different loss terms. Next, we compared meta-testing with fine-tuning and different numbers of pseudo pairs included in the support set, shown in Fig.~\ref{fig:sup_num}.


    
    




\noindent\textbf{Analyses of the adopted losses in reblurring model:} 
Our approach adopts the adversarial and cycle losses to update the reblurring model, as in Eq.~\ref{reblur_update}. 
%
In Table~\ref{table:reblur_loss}, we analyze the effects of using these losses for deblurring on the DVD test set. 
%
%
``Baseline" here denotes the performance using the original models. 
%
``Cycle loss" and ``GAN loss" mean updating the reblurring model with either the cycle loss or adversarial loss used in the inner update. 
As observed, using either the cycle loss or GAN loss improves the performance, and adopting both losses leads to more performance gain.
%

\setlength{\tabcolsep}{6pt} 
\renewcommand{\arraystretch}{1} 
\renewcommand{\cellalign}{cc}

\begin{table}[t!]
\centering
\caption{Comparisons among different losses used in the reblurring model on DVD dataset.}
\begin{tabular}{c | c c c c}
\toprule

    & \bf{Baseline} & \bf{Cycle loss} & \bf{GAN loss} & \bf{GAN+Cycle loss}\\
    \hline
    \bf{MIMOUnet+} & 29.43 & 29.58 & 29.62 & 29.70 (\red{+0.27}) \\
    \hline
    \bf{MPRNet} & 29.68 & 29.83 & 30.02 & 30.04 (\red{+0.36}) \\
    \hline
    \bf{Restormer} & 29.67 & 29.81 & 29.90 & 30.01 (\red{+0.34}) \\
    \hline
    \bf{CDVD-TSP} & 30.83 & 30.94 & 30.94 & 30.97 (\red{+0.14}) \\
    
\bottomrule
\end{tabular}
\label{table:reblur_loss}
\end{table}

\noindent\textbf{Meta-testing and number of pseudo pairs used:} 
%
%
%
Fig.~\ref{fig:sup_num} compares our meta-transferring scheme with fine-tuning, which means directly fine-tuning the models with the pseudo pairs instead of meta-learning. It shows that adopting our scheme outperforms merely fine-tuning for all the state-of-the-art methods on the benchmark datasets. We can also observe that including more pseudo pairs in the support set can increase the performance gain but with diminishing marginal benefits since selecting more pseudo-sharp patches may include ones with much blur and harm the performance. In our experiment, $M$ is set to 20 for DVD and RealBlur-J and 10 for REDS. We use more pairs in DVD~\cite{su2017deep} and RealBlur-J~\cite{rim2020real} since they have less blur for finding good pseudo-sharp patches to help test-time adaptation. 



\setlength{\tabcolsep}{2pt} 
\renewcommand{\arraystretch}{0.1} 

\noindent\textbf{Computation cost and limitation:}
For meta-testing, each video needs to run the inner update for ten iterations on the REDS dataset and twenty on the DVD and RealBlur-J datasets.
Take MIMO-UNet+ as an example. We need additional 0.1 seconds on average for inferencing one image.
A limitation of the proposed scheme is that when we deal with a strongly blurred video containing few relative-sharp patches,  the reblur-deblur meta-transferring process would be less effective, possibly leading to poor performance.



\section{Conclusion}
\label{sec:conclu}
This paper proposed a novel reblur-deblur meta-transferring scheme to facilitate meta-learning without ground truth at inference time. We combine a reblurring model to generate pseudo-blurred patches from selected pseudo-sharp patches as the support set, enabling meta-learning to update deblurring models during testing. Extensive experiments have shown that our method can improve existing deblurring models on benchmark datasets, including DVD, REDS, and RealBlur-J.
%
%

\newpage

\section*{Acknowledgements}
This work was supported in part by National Science and Technology Council (NSTC) under grants 111-2628-E-A49-025-MY3, 109-2221-E-009-113-MY3, 110-2634-F-006-022, 110-2634-F-002-050, 111-2634-F-007-002, 111-2221-E-004-010, and 110-2622-E-004-001. This work was funded in part by Qualcomm Technologies, Inc., through a Taiwan University Research Collaboration Project, under Grant NAT-487844 and MediaTek. We thank to National Center for High-performance Computing (NCHC) of National Applied Research Laboratories (NARLabs) in Taiwan for providing computational and storage resources.

\bibliography{egbib}

\begin{thebibliography}{39}
\providecommand{\natexlab}[1]{#1}
\providecommand{\url}[1]{\texttt{#1}}
\expandafter\ifx\csname urlstyle\endcsname\relax
  \providecommand{\doi}[1]{doi: #1}\else
  \providecommand{\doi}{doi: \begingroup \urlstyle{rm}\Url}\fi

\bibitem[Argaw et~al.(2021)Argaw, Kim, Rameau, and Kweon]{argaw2021motion}
Dawit~Mureja Argaw, Junsik Kim, Francois Rameau, and In~So Kweon.
\newblock Motion-blurred video interpolation and extrapolation.
\newblock In \emph{AAAI Conference on Artificial Intelligence}, 2021.

\bibitem[Chen et~al.(2022)Chen, Panda, and Fan]{chen2021regionvit}
Chun-Fu Chen, Rameswar Panda, and Quanfu Fan.
\newblock Regionvit: Regional-to-local attention for vision transformers.
\newblock \emph{Proc. Int'l Conf. Learning Representations}, 2022.

\bibitem[Chen et~al.(2018)Chen, Gu, Gallo, Liu, Veeraraghavan, and
  Kautz]{chen2018reblur2deblur}
Huaijin Chen, Jinwei Gu, Orazio Gallo, Ming-Yu Liu, Ashok Veeraraghavan, and
  Jan Kautz.
\newblock Reblur2deblur: Deblurring videos via self-supervised learning.
\newblock In \emph{IEEE International Conference on Computational Photography},
  2018.

\bibitem[Chen et~al.(2019)Chen, Fang, Wang, and Zhang]{Chen_2019_CVPR}
Liang Chen, Faming Fang, Tingting Wang, and Guixu Zhang.
\newblock Blind image deblurring with local maximum gradient prior.
\newblock In \emph{Proceedings of the IEEE/CVF Conference on Computer Vision
  and Pattern Recognition}, 2019.

\bibitem[Chi et~al.(2021)Chi, Wang, Yu, and Tang]{chi2021test}
Zhixiang Chi, Yang Wang, Yuanhao Yu, and Jin Tang.
\newblock Test-time fast adaptation for dynamic scene deblurring via
  meta-auxiliary learning.
\newblock In \emph{Proceedings of the IEEE/CVF Conference on Computer Vision
  and Pattern Recognition}, 2021.

\bibitem[Cho et~al.(2021)Cho, Ji, Hong, Jung, and Ko]{cho2021rethinking}
Sung-Jin Cho, Seo-Won Ji, Jun-Pyo Hong, Seung-Won Jung, and Sung-Jea Ko.
\newblock Rethinking coarse-to-fine approach in single image deblurring.
\newblock In \emph{Proceedings of the IEEE/CVF International Conference on
  Computer Vision}, 2021.

\bibitem[Cho and Lee(2009)]{Cho_2009_ACM}
Sunghyun Cho and Seungyong Lee.
\newblock Fast motion deblurring.
\newblock In \emph{ACM Trans. on Graphics}, 2009.

\bibitem[Deng et~al.(2021)Deng, Ren, Yan, Wang, Song, and Cao]{deng2021multi}
Senyou Deng, Wenqi Ren, Yanyang Yan, Tao Wang, Fenglong Song, and Xiaochun Cao.
\newblock Multi-scale separable network for ultra-high-definition video
  deblurring.
\newblock In \emph{Proceedings of the IEEE/CVF International Conference on
  Computer Vision}, 2021.

\bibitem[Finn et~al.(2017)Finn, Abbeel, and Levine]{finn2017model}
Chelsea Finn, Pieter Abbeel, and Sergey Levine.
\newblock Model-agnostic meta-learning for fast adaptation of deep networks.
\newblock In \emph{Proc. Int'l Conf. Machine Learning}, 2017.

\bibitem[Gulrajani et~al.(2017)Gulrajani, Ahmed, Arjovsky, Dumoulin, and
  Courville]{Wgan}
Ishaan Gulrajani, Faruk Ahmed, Martin Arjovsky, Vincent Dumoulin, and Aaron~C.
  Courville.
\newblock Improved training of wasserstein gans.
\newblock In \emph{Proc. Neural Information Processing Systems}, 2017.

\bibitem[Kingma and Ba(2015)]{kingma2014adam}
Diederik~P Kingma and Jimmy Ba.
\newblock Adam: A method for stochastic optimization.
\newblock \emph{Proc. Int'l Conf. Learning Representations}, 2015.

\bibitem[Kupyn et~al.(2019)Kupyn, Martyniuk, Wu, and Wang]{kupyn2019deblurgan}
Orest Kupyn, Tetiana Martyniuk, Junru Wu, and Zhangyang Wang.
\newblock Deblurgan-v2: Deblurring (orders-of-magnitude) faster and better.
\newblock In \emph{Proceedings of the IEEE/CVF International Conference on
  Computer Vision}, 2019.

\bibitem[Lai et~al.(2017)Lai, Huang, Ahuja, and Yang]{Lai_2017_CVPR}
Wei-Sheng Lai, Jia-Bin Huang, Narendra Ahuja, and Ming-Hsuan Yang.
\newblock Deep laplacian pyramid networks for fast and accurate
  super-resolution.
\newblock In \emph{Proceedings of the IEEE/CVF Conference on Computer Vision
  and Pattern Recognition}, 2017.

\bibitem[Li et~al.(2021)Li, Xu, Zhang, Yu, Zhong, Ren, Suominen, and
  Li]{li2021arvo}
Dongxu Li, Chenchen Xu, Kaihao Zhang, Xin Yu, Yiran Zhong, Wenqi Ren, Hanna
  Suominen, and Hongdong Li.
\newblock Arvo: Learning all-range volumetric correspondence for video
  deblurring.
\newblock In \emph{Proceedings of the IEEE/CVF Conference on Computer Vision
  and Pattern Recognition}, 2021.

\bibitem[Lin et~al.(2020)Lin, Zhang, Pan, Jiang, Zou, Wang, Chen, and
  Ren]{lin2020learning}
Songnan Lin, Jiawei Zhang, Jinshan Pan, Zhe Jiang, Dongqing Zou, Yongtian Wang,
  Jing Chen, and Jimmy Ren.
\newblock Learning event-driven video deblurring and interpolation.
\newblock In \emph{European Conference on Computer Vision}, 2020.

\bibitem[Nah et~al.(2017)Nah, Hyun~Kim, and Mu~Lee]{nah2017deep}
Seungjun Nah, Tae Hyun~Kim, and Kyoung Mu~Lee.
\newblock Deep multi-scale convolutional neural network for dynamic scene
  deblurring.
\newblock In \emph{Proceedings of the IEEE conference on computer vision and
  pattern recognition}, 2017.

\bibitem[Nah et~al.(2019{\natexlab{a}})Nah, Baik, Hong, Moon, Son, Timofte, and
  Mu~Lee]{nah2019ntire}
Seungjun Nah, Sungyong Baik, Seokil Hong, Gyeongsik Moon, Sanghyun Son, Radu
  Timofte, and Kyoung Mu~Lee.
\newblock Ntire 2019 challenge on video deblurring and super-resolution:
  Dataset and study.
\newblock In \emph{Proceedings of the IEEE/CVF Conference on Computer Vision
  and Pattern Recognition Workshops}, 2019{\natexlab{a}}.

\bibitem[Nah et~al.(2019{\natexlab{b}})Nah, Son, and Lee]{nah2019recurrent}
Seungjun Nah, Sanghyun Son, and Kyoung~Mu Lee.
\newblock Recurrent neural networks with intra-frame iterations for video
  deblurring.
\newblock In \emph{Proceedings of the IEEE/CVF Conference on Computer Vision
  and Pattern Recognition}, 2019{\natexlab{b}}.

\bibitem[Nah et~al.(2022)Nah, Son, Lee, and Lee]{nah2021clean}
Seungjun Nah, Sanghyun Son, Jaerin Lee, and Kyoung~Mu Lee.
\newblock Clean images are hard to reblur: A new clue for deblurring.
\newblock \emph{Proc. Int'l Conf. Learning Representations}, 2022.

\bibitem[Pan et~al.(2018)Pan, Sun, Pfister, and Yang]{Pan_2016_cvpr}
Jinshan Pan, Deqing Sun, Hanspeter Pfister, and Ming-Hsuan Yang.
\newblock Deblurring images via dark channel prior.
\newblock In \emph{Proceedings of the IEEE/CVF Conference on Computer Vision
  and Pattern Recognition}, 2018.

\bibitem[Pan et~al.(2020)Pan, Bai, and Tang]{pan2020cascaded}
Jinshan Pan, Haoran Bai, and Jinhui Tang.
\newblock Cascaded deep video deblurring using temporal sharpness prior.
\newblock In \emph{Proceedings of the IEEE/CVF Conference on Computer Vision
  and Pattern Recognition}, 2020.

\bibitem[Park et~al.(2020{\natexlab{a}})Park, Kang, and Chun]{park2020blur}
Dongwon Park, Dong~Un Kang, and Se~Young Chun.
\newblock Blur more to deblur better: Multi-blur2deblur for efficient video
  deblurring.
\newblock \emph{arXiv preprint arXiv:2012.12507}, 2020{\natexlab{a}}.

\bibitem[Park et~al.(2020{\natexlab{b}})Park, Kang, Kim, and
  Chun]{park2020multi}
Dongwon Park, Dong~Un Kang, Jisoo Kim, and Se~Young Chun.
\newblock Multi-temporal recurrent neural networks for progressive non-uniform
  single image deblurring with incremental temporal training.
\newblock In \emph{European Conference on Computer Vision}, 2020{\natexlab{b}}.

\bibitem[Park et~al.(2020{\natexlab{c}})Park, Yoo, Cho, Kim, and
  Kim]{park2020fast}
Seobin Park, Jinsu Yoo, Donghyeon Cho, Jiwon Kim, and Tae~Hyun Kim.
\newblock Fast adaptation to super-resolution networks via meta-learning.
\newblock In \emph{European Conference on Computer Vision}, 2020{\natexlab{c}}.

\bibitem[Rim et~al.(2020)Rim, Lee, Won, and Cho]{rim2020real}
Jaesung Rim, Haeyun Lee, Jucheol Won, and Sunghyun Cho.
\newblock Real-world blur dataset for learning and benchmarking deblurring
  algorithms.
\newblock In \emph{European Conference on Computer Vision}, 2020.

\bibitem[Shen et~al.(2020)Shen, Bao, Zhai, Chen, Min, and Gao]{shen2020blurry}
Wang Shen, Wenbo Bao, Guangtao Zhai, Li~Chen, Xiongkuo Min, and Zhiyong Gao.
\newblock Blurry video frame interpolation.
\newblock In \emph{Proceedings of the IEEE/CVF Conference on Computer Vision
  and Pattern Recognition}, 2020.

\bibitem[Soh et~al.(2020)Soh, Cho, and Cho]{soh2020meta}
Jae~Woong Soh, Sunwoo Cho, and Nam~Ik Cho.
\newblock Meta-transfer learning for zero-shot super-resolution.
\newblock In \emph{Proceedings of the IEEE/CVF Conference on Computer Vision
  and Pattern Recognition}, 2020.

\bibitem[Su et~al.(2017)Su, Delbracio, Wang, Sapiro, Heidrich, and
  Wang]{su2017deep}
Shuochen Su, Mauricio Delbracio, Jue Wang, Guillermo Sapiro, Wolfgang Heidrich,
  and Oliver Wang.
\newblock Deep video deblurring for hand-held cameras.
\newblock In \emph{Proceedings of the IEEE Conference on Computer Vision and
  Pattern Recognition}, 2017.

\bibitem[Suin and Rajagopalan(2021)]{suin2021gated}
Maitreya Suin and AN~Rajagopalan.
\newblock Gated spatio-temporal attention-guided video deblurring.
\newblock In \emph{Proceedings of the IEEE/CVF Conference on Computer Vision
  and Pattern Recognition}, 2021.

\bibitem[Tsai* et~al.(2021)Tsai*, Peng*, Lin, Tsai, and Lin]{Tsai2022BANet}
Fu-Jen Tsai*, Yan-Tsung Peng*, Yen-Yu Lin, Chung-Chi Tsai, and Chia-Wen Lin.
\newblock Banet: Blur-aware attention networks for dynamic scene deblurring.
\newblock In \emph{arXiv preprint arXiv:2101.07518}, 2021.

\bibitem[Tsai et~al.(2022)Tsai, Peng, Lin, Tsai, and Lin]{Tsai2022Stripformer}
Fu-Jen Tsai, Yan-Tsung Peng, Yen-Yu Lin, Chung-Chi Tsai, and Chia-Wen Lin.
\newblock Stripformer: Strip transformer for fast image deblurring.
\newblock In \emph{European Conference on Computer Vision}, 2022.

\bibitem[Wang et~al.(2022)Wang, Cun, Bao, and Liu]{wang2021uformer}
Zhendong Wang, Xiaodong Cun, Jianmin Bao, and Jianzhuang Liu.
\newblock Uformer: A general u-shaped transformer for image restoration.
\newblock \emph{Proceedings of the IEEE/CVF Conference on Computer Vision and
  Pattern Recognition}, 2022.

\bibitem[Yan et~al.(2017)Yan, Ren, Guo, Wang, and Cao]{8100221}
Yanyang Yan, Wenqi Ren, Yuanfang Guo, Rui Wang, and Xiaochun Cao.
\newblock Image deblurring via extreme channels prior.
\newblock In \emph{Proceedings of the IEEE/CVF Conference on Computer Vision
  and Pattern Recognition}, 2017.

\bibitem[Zamir et~al.(2021)Zamir, Arora, Khan, Hayat, Khan, Yang, and
  Shao]{Zamir2021MPRNet}
Syed~Waqas Zamir, Aditya Arora, Salman Khan, Munawar Hayat, Fahad~Shahbaz Khan,
  Ming-Hsuan Yang, and Ling Shao.
\newblock Multi-stage progressive image restoration.
\newblock In \emph{Proceedings of the IEEE/CVF Winter Conference on
  Applications of Computer Vision}, 2021.

\bibitem[Zamir et~al.(2022)Zamir, Arora, Khan, Hayat, Khan, and
  Yang]{zamir2021restormer}
Syed~Waqas Zamir, Aditya Arora, Salman Khan, Munawar Hayat, Fahad~Shahbaz Khan,
  and Ming-Hsuan Yang.
\newblock Restormer: Efficient transformer for high-resolution image
  restoration.
\newblock In \emph{Proceedings of the IEEE/CVF Conference on Computer Vision
  and Pattern Recognition}, 2022.

\bibitem[Zhang et~al.(2019)Zhang, Dai, Li, and Koniusz]{zhang2019deep}
Hongguang Zhang, Yuchao Dai, Hongdong Li, and Piotr Koniusz.
\newblock Deep stacked hierarchical multi-patch network for image deblurring.
\newblock In \emph{Proceedings of the IEEE/CVF Conference on Computer Vision
  and Pattern Recognition}, 2019.

\bibitem[Zhang et~al.(2020)Zhang, Luo, Zhong, Ma, Stenger, Liu, and
  Li]{zhang2020deblurring}
Kaihao Zhang, Wenhan Luo, Yiran Zhong, Lin Ma, Bjorn Stenger, Wei Liu, and
  Hongdong Li.
\newblock Deblurring by realistic blurring.
\newblock In \emph{Proceedings of the IEEE/CVF Conference on Computer Vision
  and Pattern Recognition}, 2020.

\bibitem[Zhong et~al.(2020)Zhong, Gao, Zheng, and Zheng]{zhong2020efficient}
Zhihang Zhong, Ye~Gao, Yinqiang Zheng, and Bo~Zheng.
\newblock Efficient spatio-temporal recurrent neural network for video
  deblurring.
\newblock In \emph{European Conference on Computer Vision}, 2020.

\bibitem[Zhou et~al.(2019)Zhou, Zhang, Pan, Xie, Zuo, and Ren]{zhou2019spatio}
Shangchen Zhou, Jiawei Zhang, Jinshan Pan, Haozhe Xie, Wangmeng Zuo, and Jimmy
  Ren.
\newblock Spatio-temporal filter adaptive network for video deblurring.
\newblock In \emph{Proceedings of the IEEE/CVF International Conference on
  Computer Vision}, 2019.

\end{thebibliography}
\end{document}